\documentclass[letterpaper, 10 pt, conference]{ieeeconf}  % Comment this line out
\usepackage{amsmath} 
\usepackage{amsfonts}
\usepackage{color}
\usepackage{graphicx}
\IEEEoverridecommandlockouts                              % This command is only
 \title{\LARGE \bf Integrated Pre-Processing for Bayesian Nonlinear System Identification \\ with Gaussian Processes}

\author{Roger Frigola and Carl Edward Rasmussen% <-this % stops a space
\thanks{The authors are with the Department of Engineering, University of Cambridge, UK, {\tt \{rf342,cer54\}@cam.ac.uk}}
\thanks{All the results presented in this paper can be reproduced without any tuning by using the code for GP-FNARX available from the first author's website: \texttt{http://mlg.eng.cam.ac.uk/roger}}%
}

\begin{document}

\maketitle
\thispagestyle{empty}
\pagestyle{empty}

%%%%%%%%%%%%%%%%%%%%%%%%%%%%%%%%%%%%%%%%%%%%%%%%%%%%%%%%%%%%%%%%%%%%%%%%%%%%%%%%
\begin{abstract}

%We present a methodology for (Bayesian) nonlinear system identification that automatically performs the data pre-processing step that is usually carried out manually. We interleave data pre-processing with system identification using a Bayesian auto-regressive model based on Gaussian processes. Noise modeling is accomplished in a completely data-driven fashion. %very close to the Nonlinear Auto-Regressive with eXogenous inputs (NARX) model based on Bayesian nonparametric inference methods.

We introduce GP-FNARX: a new model for nonlinear system identification based on a nonlinear autoregressive exogenous model (NARX) with filtered regressors (F) where the nonlinear regression problem is tackled using sparse Gaussian processes (GP). We integrate data pre-processing with system identification into a fully automated procedure that goes from raw data to an identified model. Both pre-processing parameters and GP hyper-parameters are tuned by maximizing the marginal likelihood of the probabilistic model. We obtain a Bayesian model of the system's dynamics which is able to report its uncertainty in regions where the data is scarce. The automated approach, the modeling of uncertainty and its relatively low computational cost make of GP-FNARX a good candidate for applications in robotics and adaptive control. 

\end{abstract}

%%%%%%%%%%%%%%%%%%%%%%%%%%%%%%%%%%%%%%%%%%%%%%%%%%%%%%%%%%%%%%%%%%%%%%%%%%%%%%%%
\section{Introduction}

System identification consists in building mathematical models of dynamical systems based on observed input and output signals. Those signals are often contaminated by noise and do not necessarily explore the whole operating range of the system. Yet, system identification methods are required to find a model that faithfully explains the observed data and has graceful generalization capabilities. % outside of the explored operating range.

In this paper we will be particularly concerned with identifying models of nonlinear systems when only a limited amount of data are available. Data could be scarce due to the high cost of experimentation, vast operating range of the system, presence of closed loop control or many other reasons. As a consequence, we will require a model that can accurately describe the system wherever data is present but which can also report its own ignorance in regions where no data can support its claims. This naturally leads to the use of a Bayesian framework for the identification of the system's dynamics \cite{Peterka1981}.

A very common and general representation of a dynamical system is in terms of a \emph{state space model}. The future behavior of the system is fully described by its current state $z$ and external input $u$. In statistical terms, future states are conditionally independent of past states and inputs given the current ones
\begin{equation}
p(z_{t+1}|z_t,z_{t-1},...,u_t,u_{t-1},...) = p(z_{t+1}|z_t,u_t).
\end{equation}
A system with this characteristic is said to have the Markov property.

The state of a system is often not perfectly observed. An observation model is usually defined as a probability density of an observation $y_t$ conditioned on the state at the same time step: $p(y_t|z_t)$. In the particular case where a nonlinear state space model has additive independent Gaussian noise in both the state transition and the observation, the model can be presented as
\begin{align}\label{ssm1}
z_{t+1} & =  f( z_t , u_t) + \delta_t ,  &&\delta_t \sim \mathcal{N}(0,Q)\\
y_t & = g( z_t) + \varepsilon_t, && \varepsilon_t \sim \mathcal{N}(0,R). \label{ssm2}
\end{align}
For time-invariant systems, $f$ and $g$ represent fixed deterministic functions although both the state transition and the observation model are stochastic difference equations (\ref{ssm1},\ref{ssm2}).

In statistical terms, the system identification task can be described as finding the state transition probability conditioned on the observed inputs and outputs
\begin{equation}
p(z_{t+1}|z_t,u_t,u_{1:N},y_{1:N}),\quad t>N.
\end{equation}
The fact that the states are not observed makes this distribution exceedingly difficult to compute. In particular, if one wants to obtain a model that is able to report its own uncertainty (i.e. a Bayesian model), there exist no closed form solutions even for the simplest linear-Gaussian state space models \cite{Barber2006,Wills2012}.

%Several approaches to the problem have been made (cite a couple \cite{Schoen2011}) but efficient learning of the system dynamics for nonlinear state space models is still a very hot research topic. One can make certain approximations to the problem ML approach with EM using particle smoothing and a parametric model for the dynamics \cite{Schoen2011} 
%Is EM actually the Bayesian way to do it? or the is it a 

A popular alternative to state space models can be found in autoregressive exogenous (ARX) models \cite{Ljung1999}. These models represent the system based only on observable quantities: the inputs and the outputs. A nonlinear ARX (NARX) model with Gaussian innovations provides an estimate of the next output based on a finite amount of previous inputs and outputs
\begin{equation}\label{narx}
y_t  =  f( y_{t-1}, y_{t-2}, ... , u_{t-1}, ...) + \delta_t,  \quad \delta_t \sim \mathcal{N}(0,Q).
\end{equation}
In contrast with state space models, there are no latent states and the model does not have the Markov property. In practical terms, this representation based only on observables makes it possible to sidestep many of the problems associated with system identification when latent states are present.

An important drawback of autoregressive models is that they do not account for \emph{observation noise}. If we consider equation \eqref{narx} in a generative manner by assuming a known $f$ and sequentially generate new values of $y_t, y_{t+1}, ...$, it is apparent that all randomness injected through the innovation $\delta$ has an influence on the future trajectory of the system. In other words, the innovation can be considered as a form of \emph{process noise}. The model does not accommodate for any kind of observation noise such as the one described in equation~\eqref{ssm2} for the case of state space models.

A common way to deal with observation noise is to pre-process the measured input and output data to remove as much noise as possible before identifying the NARX model \cite{Ljung1999}. For instance, low-pass filtering the signals can remove a significant component of their noise. However, for optimal performance, the filter needs to be tuned to the particular characteristics of the noise.

We propose an automated approach which simultaneously performs filter tuning and learns a Bayesian model of the system's dynamics. Noise modeling is performed in a completely data-driven fashion by optimizing the \emph{marginal likelihood} of the probabilistic regression model. 

%Our method is conceptually close to state space models in the sense that we consider that observations are corrupted by noise and that there are latent variables inaccessible to us. However, our method is simpler since the latent space is the space of observables.

% my method (I call it AR for the non-Markov structure and because we work in the space of observations but it's closer to the state space approach in the sense that it has latent variables that we don't treat in the most rigorous manner but in one that avoids the coupled inference + learning problem which requires EM...)

In section \ref{sec:related} we compare our approach to related work. Section \ref{prelim} presents an overview of the field of Bayesian system identification together with a brief introduction to Gaussian processes. Then, in section \ref{model}, we describe our approach to system identification integrating data pre-processing with learning the system's dynamics. We show experimental results in section \ref{exp} and, finally, in section \ref{conc} we make concluding remarks.

%%%%%%%%%%%%%%%%%%%%%%%%%%%%%%%%%%%%%%%%%%%%%%%%%%%%%%%%%%%%%%%%%%%%%%%%%%%%%%%%
\section{Related Work}\label{sec:related}

Kocijan et al. \cite{Kocijan2005} presented a NARX model where the nonlinearity is captured using Gaussian process regression. More recently Gutjahr et al. \cite{Gutjahr2012} have proposed the use of FITC in order to make computation more efficient in this kind of models. Our model follows this approach but also considers that the input/output signals may be contaminated by noise which presents an \emph{errors-in-variables} regression problem.

We mitigate the errors-in-variables problem by incorporating into system identification the signal pre-processing step that is usually carried out manually \cite{Ljung1999,Ljung2012}. For instance, a portion of the noise can be removed by low-pass filtering the signals in the time domain. 

More sophisticated approaches combining smoothing and system identification for state space models have the potential to give better results if one is ready to accept their additional complexity \cite{TurDeiRas10,Schoen2011}. In contrast, our method needs to be seen as a pragmatic approach that attempts to offer a better solution than standard NARX models with a simple procedure and a limited amount of computational effort. Thus, it is particularly well suited to large datasets.

%by putting a time series problem in the form of regression we lose some of the temporal couplings that are useful to eliminate observation noise (how can i make this claim more specific? example, intuition...)

%\cite{Ljung2008} Issues in Identification of Nonlinear Models
%\cite{Ljung2011} machine learning in system identification
%\cite{Sjöberg1995} state NARX regressors should be the first method we attempt for nonlinear sysid p. 1721

%Rivera 1992 Control-relevant prefiltering: a systematic design approach and case study "some caution must be exercised when applied to FIR and low-order ARX estimation"

%\cite{Espinoza2005} ???

%\cite{Ahn2010} fuzzy, robotic arm, 2010, quite well cited wrt IEEE NARX papers

%%%%%%%%%%%%%%%%%%%%%%%%%%%%%%%%%%%%%%%%%%%%%%%%%%%%%%%%%%%%%%%%%%%%%%%%%%%%%%%%
\section{Background Theory}\label{prelim}
\subsection{Bayesian System Identification}\label{bsi}

The goal of Bayesian system identification is to obtain models of dynamical systems that quantify the degree of uncertainty in the system's dynamics \cite{Peterka1981}. Modeling uncertainty can be particularly useful in applications such as robotics or adaptive control where decisions about future control actions will typically depend on the confidence that the agent has about the system dynamics.

Bayesian system identification relies on the use of Bayesian statistics and hence interprets probabilities as degrees of belief. Such an approach to probabilistic modeling has been very successful in the fields of machine learning and artificial intelligence \cite{Bishop2006}. We will borrow heavily from advances in those fields to model uncertainty in a principled manner. In particular, we will solve the nonlinear regression task from a NARX model with Bayesian nonlinear regression implemented with Gaussian processes.

% however not fully developed due to computational complexity. Isermann or Nelles books (or even Ljung, for that matter) do not develop Bayesian inference in detail. Possibly because it is not until recently that methodological and computational advances have Bayesian methods cheap enough.

\subsection{Gaussian Process Regression}\label{gps}

Gaussian processes are stochastic processes that have proven very successful at the task of nonlinear regression. Their unbounded flexibility makes them ideal whenever it is hard to specify a parametric form for an unknown nonlinear function.

Formally, a Gaussian process is a collection of random variables, \emph{any finite number of which have a joint Gaussian distribution} \cite{RasWil06}. We say that a random function $f(x) \in \mathbb{R}$, with $x \in \mathbb{R}^n$, follows a Gaussian process and write it as
\begin{equation}
f(x) \sim \mathcal{GP}(m(x),k(x,x'))
\end{equation}
when all values of $f$ at any locations $x$ are jointly normally distributed. $m(x)$ and $k(x,x')$ represent the mean function and the covariance function respectively. Together, they completely specify the Gaussian process and can be parametrized by a, typically small, set of hyper-parameters. Those hyper-parameters define the kind of functions that one expects to see. For instance, they control the degree of smoothness or periodicity but do not constrain the function to have any predefined shape.

In Gaussian process regression, a likelihood function $p(y|f)$ relates the vector of real-valued observations $y=(y_1,...,y_N)$ with the latent function $f$. In the particular case where observations are contaminated with additive Gaussian i.i.d. noise 
\begin{equation}\label{eq:lik}
p(y_i|f_i) = \mathcal{N}(y_i|f_i,\sigma^2),
\end{equation}
there exists a convenient closed-form solution for the posterior distribution over functions, i.e. the distribution over the unknown function $f(x)$ after incorporating all the observed data. The \emph{posterior distribution} of the function at any new location $x_*$ is described by
\begin{equation}
p(f_*|x_*,y,X) = \mathcal{N}(\bar{f}_*,\mathrm{cov}(f_*))
\end{equation}
where $X=\{x_1, ..., x_N\}$ is the set of regressor vectors.  In the common case where $m(x) = 0$ we obtain
\begin{align}
\bar{f}_* & =  K(X_*,X) [K(X,X) + \sigma^2 I]^{-1} y, \label{eq:gpmean}\\
\mathrm{cov}(f_*) & = K(X_*,X_*) - \nonumber \\ 
& \quad K(X_*,X) [K(X,X) + \sigma^2 I]^{-1} K(X,X_*). \label{eq:gpcov}
\end{align}
where the covariance matrices $K$ are constructed by applying the covariance function to all elements in each of the sets that they take as arguments:
\begin{equation}\label{eq:covmat}
K(A,B)_{ij} = k(a_i,b_j).
\end{equation}

\begin{figure}[t]
\centering
\includegraphics[width=6cm]{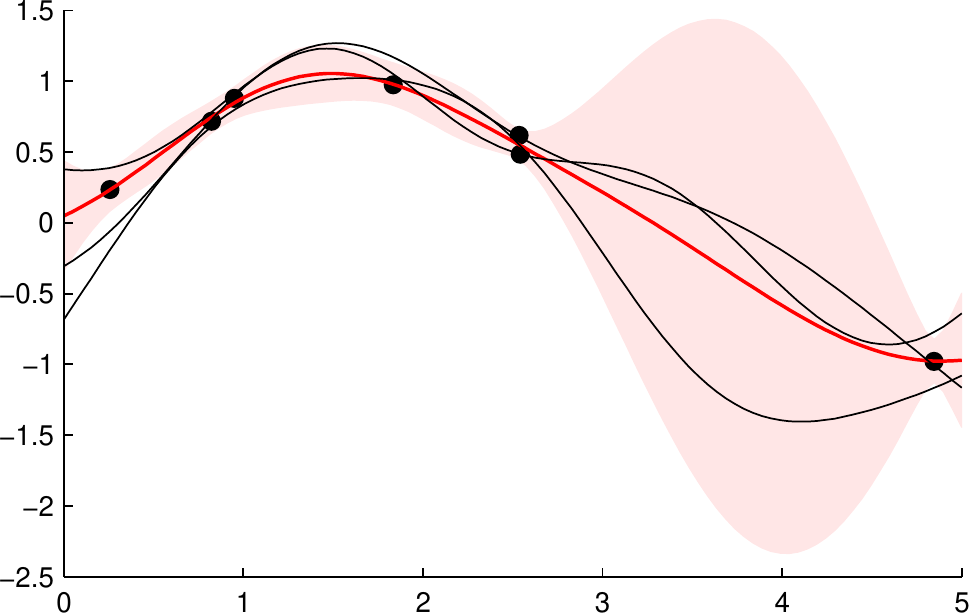}
\vspace{0in}
\caption{Gaussian process regression. Black dots represent the data. The red line is the mean of the Gaussian process prediction and the area shaded in red covers the 95\% confidence region. The three black lines are random functions drawn from the posterior that could have generated the data.}\label{fig:demogp}
\end{figure}

In Figure \ref{fig:demogp} we show an example of Gaussian process regression. Given a set of seven noisy data points, the goal is to infer the latent function that generated them. We do not assume any parametric form for this latent function. Our assumption is that the latent function was a draw from a Gaussian process with zero mean and a squared exponential covariance function \cite{RasWil06}. To obtain the posterior distribution over the latent function we use equations (\ref{eq:gpmean}) and (\ref{eq:gpcov}). Note how the confidence region widens in areas that are far from the data points.

\subsection{Model Selection for Gaussian Processes}\label{sec:modelselec}

In the Gaussian process framework, model selection refers to using data to select the functional form for the mean and the covariance functions as well as selecting the values for their hyper-parameters. In addition, the likelihood function may also have some hyper-parameters, such as the noise variance $\sigma^2$ in equation (\ref{eq:lik}). For convenience we include the likelihood hyper-parameters into the vector $\theta$.

For Gaussian process models, it is very attractive, from a computational and theoretical point of view, to use Bayesian model selection procedures. In particular, maximization of the \emph{marginal likelihood} of the Gaussian process model has proven to be very effective \cite{RasWil06}. The objective is to find
\begin{equation}
\theta_{\rm Opt} = \operatorname*{arg\,max}_\theta \ p(y|X,\theta)
\end{equation}
where
\begin{equation}
p(y|X,\theta) = \int p(y|f,\theta)\ p(f|X,\theta)\ df
\end{equation}
is the marginal likelihood. The first term inside the integral is the likelihood function and the second term is a normal distribution that can be obtained directly from the mean and covariance functions that specify the Gaussian process. For the particular case of Gaussian processes with zero mean and a Gaussian likelihood, such as in eq. (\ref{eq:lik}), we obtain:
\begin{equation}\label{eq:marglikgauss}
p(y|X,\theta) = \mathcal{N}(0,K_\theta(X,X) + \sigma^2 I).
\end{equation}
To avoid underflow errors, it is common to work with the logarithm of the marginal likelihood rather than the marginal likelihood itself. Since the logarithm function is strictly monotonically increasing, this transformation does not change the location of the optimum. If we take logarithms on both sides of the equation we obtain
\begin{equation}\label{eq:lml}
\log p(y|X,\theta) = - \frac{1}{2} y^\top K^{-1} y - \frac{1}{2} \log |K|  -\frac{n}{2} \log 2 \pi
\end{equation}
where $K=K_\theta(X,X) + {\sigma}^2 I$.

For optimization, it will be useful to have the derivative of the marginal likelihood with respect to any of the parameters of interest:
\begin{align}\label{eq:nlmder}
\frac{\partial}{\partial \theta_j} \log p(y|X(\omega),\theta) = &- \frac{1}{2} y^\top K^{-1} \frac{\partial K}{\partial \theta_j}  K^{-1} y \nonumber \\ &- \frac{1}{2} \mathrm{tr} (K^{-1} \frac{\partial K}{\partial \theta_j}).
\end{align}

The marginal likelihood embodies what is usually described as \emph{Bayesian Occam's razor}: an automatic balancing of model fit and model complexity \cite{RasWil06}. In contrast with maximum likelihood estimation, the marginal likelihood is found by \emph{integrating over the latent values of the function} $f(x)$. 

The computational complexity of the marginal likelihood in eq. (\ref{eq:marglikgauss}) is $\mathcal{O}(N^3)$ due to the inverted covariance matrix in the density of a multivariate Gaussian distribution. This cubic complexity effectively limits the approach to values of $N$ not greater than a few thousands. However, there is a mitigating factor: once $K^{-1}$ has been computed for eq. (\ref{eq:lml}), each derivative of the marginal likelihood with respect to the hyper-parameters has a computational complexity of $\mathcal{O}(N^2)$. This is a particular feature of the marginal likelihood that is useful for efficient gradient based optimization.

\subsection{Sparse Gaussian Process Regression}\label{fitc}

Several approaches have been proposed to accelerate Gaussian processes for very large data sets. Those approaches aim at reducing the particularly disadvantageous $\mathcal{O}(N^3)$ cost to train the hyper-parameters but also target a reduction of the $\mathcal{O}(N)$ cost necessary to evaluate the mean of a posterior GP (eq. \ref{eq:gpmean}) or the $\mathcal{O}(N^2)$ cost of computing its posterior covariance (eq. \ref{eq:gpcov}).

A common strategy is to introduce a new set of $M$ latent variables $\bar{f}$ at input locations $\bar{X}$ where $M < N$ \cite{QuiRas05}. The intuition behind these methods is that the new set of latent variables incorporates much of the information present in the original, large, dataset.

A particularly effective sparse GP method named FITC, or originally SPGP, was proposed by Snelson and Ghahramani \cite{SneGha06}. Due to space constraints we avoid a complete re-derivation of the FITC method and simply state its predictive distribution given a dataset $\{y,X\}$ and the locations of the latent inputs $\bar{X}$:
\begin{equation}\label{eq:fitcpred}
p(f_*|x_*,y,X,\bar{X}) = \mathcal{N}(\mu_*,\sigma_*^2),
\end{equation}
where
\begin{align}
\mu_* & = K_*^\top Q_M^{-1} K_{MN} (\Lambda + \sigma^2 I)^{-1} y \label{eq:fitcmu} \\ 
\sigma_*^2 & = K_{**} - K_*^\top (K_M^{-1} - Q_M^{-1}) K_* \label{eq:fitccov}
\end{align}
and
\begin{align}
Q_M & = K_M + K_{MN} (\Lambda + \sigma^2 I)^{-1} K_{NM} \\ 
\Lambda & = \mathrm{diag}(\lambda), \quad \lambda_n = K_{nn} - K_n^\top K_M K_n ,
\end{align}
where the matrices $K$ are constructed using the covariance function as shown in eq. (\ref{eq:covmat}) and we follow the following convention: $K_{AB} = K(X_A,X_B)$ and $K_{A} = K(X_A,X_A)$.

%%%%%%%%%%%%%%%%%%%%%%%%%%%%%%%%%%%%%%%%%%%%%%%%%%%%%%%%%%%%%%%%%%%%%%%%%%%%%%%%
\section{GP-FNARX Model}\label{model}

\subsection{Overview}

In this section, we describe the GP-FNARX model for nonlinear system identification based on a nonlinear autoregressive exogenous (NARX) model with filtered regressors (F) where the nonlinear regression problem is tackled using sparse Gaussian processes (GP). The approach integrates pre-processing (e.g. filtering) with system identification into a fully automated procedure that yields a Bayesian nonparametric model.

The key contribution of this article is the integration of the data pre-processing step with the optimization of the sparse GP hyper-parameters. Training is performed by simultaneously maximizing the \emph{marginal likelihood} of the probabilistic regression model with respect to the pre-processing parameters and the GP hyper-parameters.

Autoregressive models with exogenous inputs attempt to predict future outputs by considering them a function of past inputs and outputs to the system:
\begin{equation}
y_t  =  f( y_{t-1}, y_{t-2}, ... , u_{t-1}, ...) + \delta_t.
\end{equation}
The identification problem is then posed as a regression task. In other words, we want to infer the function $y=f(x)$ from a finite number of examples $\{y_i, x_i\}$, where $x_i = ( y_{i-1}, y_{i-2}, ... , u_{i-1}, ...)$. In our approach, we place a GP prior on the function $f(x)$ and use sparse GP regression techniques to infer it from observed data.

If the input and output signals to the dynamical system are noisy, we face an \emph{errors-in-variables} problem since the regressors $x$ are noisy. Noise in the regressors makes the regression problem particularly hard. This is one of the reasons why input signals are normally pre-processed before trying to identify a model of the system dynamics. For instance, one can carefully low-pass filter the signals to remove high-frequency noise irrelevant to the identification task at hand. Since we are looking for a method that avoids having a human in the loop, we take a data-based approach which consists in parameterizing the data pre-processing stage and optimizing its parameters jointly with the hyper-parameters of the Gaussian process regression.

We will consider \emph{any} data pre-processing function applied to the input and output signals
\begin{equation}
(\hat{y},\hat{u}) =  h(y,u,\omega)
\end{equation}
where the pre-processed signals vary \emph{smoothly} with respect to a vector of pre-processing parameters $\omega$. This smoothness condition is imposed in order to obtain a probabilistic model with a differentiable marginal likelihood.

We can rephrase the autoregressive model in terms of the pre-processed regressors:
\begin{equation}%\label{eq:narxfilt}
y_t  =  f( \hat{y}_{t-1}, \hat{y}_{t-2}, ... , \hat{u}_{t-1}, ...).
\end{equation}
Note that the left hand side term is \emph{not} pre-processed. %This is key when optimizing the marginal likelihood of the model, i.e. $p(y)$.

\subsection{Optimization of the Marginal Likelihood}

In section \ref{sec:modelselec} we have described how the marginal likelihood provides a very powerful metric to perform model selection due to its ability to automatically trade off model fit and model complexity in a principled manner. Our goal here will be to maximize the marginal likelihood of the Gaussian process regression model with respect to the signal pre-pocessing parameters and also, simultaneously, with respect to the hyper-parameters of the GP. For convenience, we introduce $\psi = (\omega,\theta)$ grouping the two kinds of parameters.

We will employ hill-climbing optimization to maximize the marginal likelihood (or, equivalently, its logarithm). For notational simplicity, we group all the pre-processed regressors into a matrix $\hat{X} = \hat{X}(y,u,\omega)$ so that the log marginal likelihood becomes
\begin{equation}
\log p(y|\hat{X}(\omega),\theta).
\end{equation}
Its derivatives with respect to the GP hyper-parameters 
\begin{align}
\frac{\partial}{\partial \theta_j} \log p(y|\hat{X}(\omega),\theta)
\end{align}
are straightforward since we typically choose differentiable covariance functions. However, the derivatives with respect to any of the pre-processing parameters
\begin{align}
\frac{\partial}{\partial \omega_k} \log p(y|\hat{X}(\omega),\theta)
\end{align}
can be more difficult to compute. From eq. (\ref{eq:nlmder}) it is apparent that $\frac{\partial K}{\partial \omega_k}$ needs to be computed.  We can write the derivative of a single element of the covariance matrix as
\begin{equation}
\frac{\partial K_{ij}}{\partial \omega_k} = \frac{\partial k(\hat{x}_i,\hat{x}_j)}{\partial \omega_k} = \frac{\partial k}{\partial \hat{x}_i} \frac{\partial \hat{x}_i}{\partial \omega_k} + \frac{\partial k}{\partial \hat{x}_j} \frac{\partial \hat{x}_j}{\partial \omega_k}
\end{equation}
where the derivatives with respect to the regressors are straightforward to compute when using smooth covariance functions. However, the derivatives of the regressors with respect to the pre-processing parameters may be hard to compute. In any case, if the pre-processing function is smooth, the derivatives $\frac{\partial \hat{x}}{\partial \omega_k}$ can be approximated numerically by finite differences at the cost of one extra evaluation of the pre-processing function per dimension of $\omega$.

\subsection{Sparse GPs for Computational Speed}

Computing the marginal likelihood for datasets with more than a few thousand points becomes computationally expensive. For larger datasets we adopt a pragmatic strategy whereby the marginal likelihood is maximized by employing only a subset of the data. 

Once the GP hyper-parameters and the data pre-processing parameters have been set, we use FITC sparse Gaussian processes \cite{QuiRas05} to obtain a model of the system dynamics that uses the \emph{whole dataset} to make predictions. Equation (\ref{eq:fitcpred}) shows the expression for FITC predictions. After all possible pre-computations, the computational complexity is $\mathcal{O}(M)$ for the mean of the predictive distribution and $\mathcal{O}(M^2)$ for its variance. The computational efficiency comes from having \emph{condensed} the original dataset with $N$ points into a smaller set of $M$ inducing points.

\begin{figure}
\hrule\vspace{4pt}
\textbf{Inputs}: $\mathcal{I}$ = \{output signals $y=(y_1,...,y_N)$, input signals $u=(u_1,...,u_N)$ , model order vector $\eta$\}
\vspace{4pt}

1.\hspace{5pt} $ \psi_0 \leftarrow$ {\sc InitialGuess$(\mathcal{I})$}

2.\hspace{5pt} $\psi_{\rm Opt} \leftarrow$ {\sc MaximizeMargLikelihood$(\psi_0,\mathcal{I})$}

3.\hspace{5pt} model $\leftarrow$ {\sc ComputeFITCPredictor$(\psi_{\rm Opt},\mathcal{I})$}

%4.\hspace{5pt}  **************
\vspace{4pt}\hrule
\caption{High-level pseudo-code for the GP-FNARX algorithm.}\label{algo}
\end{figure}

\begin{figure*}[!ht]
\centering
\includegraphics[width=16.5cm]{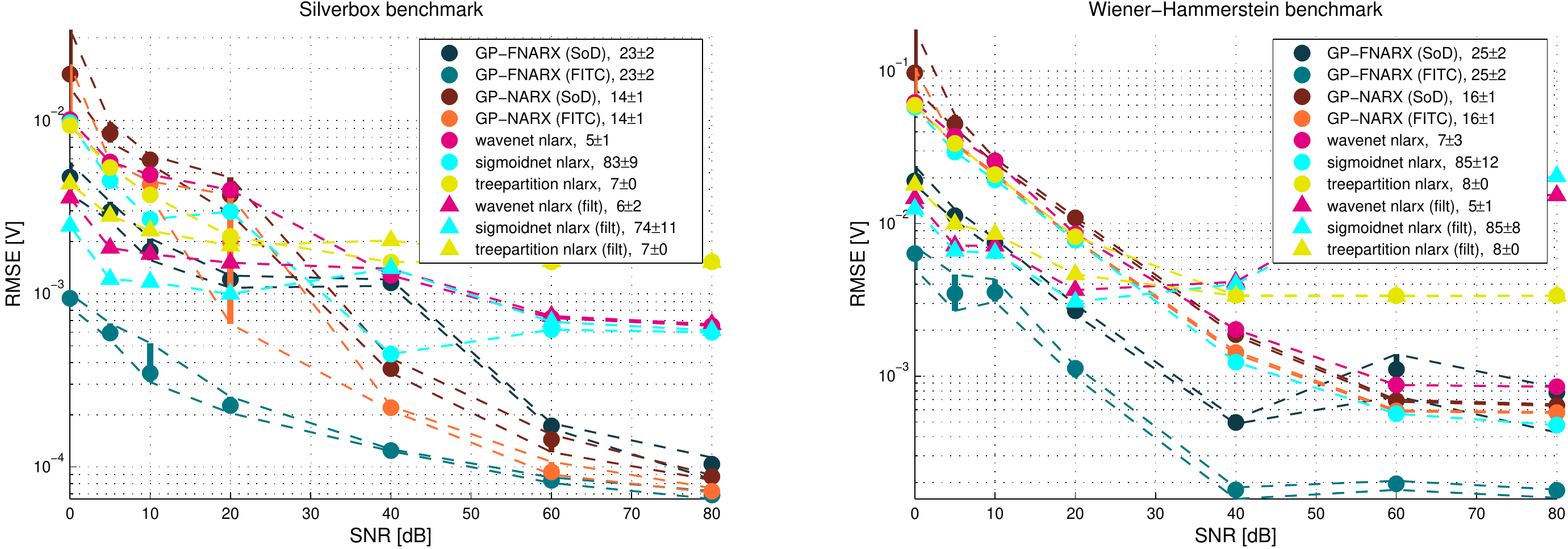}
\vspace{0in}
\caption{Root mean squared prediction error on the test set as a function of the signal to noise ratio in the output signal. The mean computation time and standard deviation for each method are displayed in the legend (in seconds). Experiments are repeated 10  times, the marker is positioned at the median value and error-bars indicate the 10-90 percentile interval.}\label{fig:prediction}
\end{figure*}

\subsection{Summary of the Proposed Methodology}

In Fig. \ref{algo} we present a high-level overview of the GP-FNARX algorithm. The first step consists in providing an initial guess for the unknown hyper-parameters. We have found that, in the absence of any problem-specific knowledge that could guide the initial guess, a successful data-based heuristic is to run a few steps of optimization of the marginal likelihood with respect to the GP hyper-parameters followed by a few steps of optimization with respect to the pre-processing parameters. By running these two steps on a subset of the data, the algorithm can rapidly hone in the right order of magnitude for the unknown parameters.

The second step consists in a straightforward joint optimization of the GP hyper-parameters and the pre-processing parameters. This step can be performed by either using a subset of the data (SoD) with a conventional GP or on the full dataset with a FITC sparse GP \cite{QuiRas05}. The FITC approach is theoretically more sound but very good and fast results can also be obtained by optimizing the marginal likelihood computed on a subset of the data.

In the third and final step, a large part of the FITC predictor equations (\ref{eq:fitcmu},\ref{eq:fitccov}) is pre-computed in order to obtain a model that will be able to provide predictive means in $\mathcal{O}(M)$ and predictive variances in $\mathcal{O}(M^2)$.

As an aside, we want to highlight that the choice of orders of the autoregressive model is not critical to the performance of the algorithm provided that two conditions are met: \emph{i)} the order is chosen to be higher than the optimal order and \emph{ii)} the automatic relevance determination (ARD) covariance function is chosen for the GP. This is due to the fact that the Bayesian Occam's razor embodied by the marginal likelihood is able to automatically disable regressors that are irrelevant to the task at hand. In our experiments we verified that adding hundreds of regressors on a problem of low order did not cause any overfitting and only represented a computation time penalty.

%%%%%%%%%%%%%%%%%%%%%%%%%%%%%%%%%%%%%%%%%%%%%%%%%%%%%%%%%%%%%%%%%%%%%%%%%%%%%%%%
\section{Experimental Evaluation}\label{exp}

In this section we present an experimental evaluation of the proposed system identification method. We have used data from two nonlinear system identification benchmarks based on electric circuits: \emph{i)} the Silverbox benchmark originating from the 2004 IFAC Symposium on Nonlinear Control Systems (NOLCOS) and \emph{ii)} the SYSID09 Wiener-Hammerstein system identification benchmark \cite{Schoukens2009} which has been the object of a special section in the November 2012 issue of the ``Control Engineering Practice" journal \cite{Hjalmarsson2012}. 

Both datasets have ${>}10^5$ data points and are corrupted by a very small amount of noise. For instance, the authors of the Wiener-Hammerstein dataset estimate its signal to noise ratio to be of 70 dB. Since we are attempting to demonstrate the ability of our method to cope with measurement noise, we will inject different amounts of synthetic i.i.d. Gaussian noise to the output signals to make the identification task more challenging. Being able to have original signals with little noise will be convenient to test the quality of the resulting models.
% access to a ground truth

Very good algorithms have been tailored to the specific benchmarks at hand. However, since our goal is to provide an automatic method for system identification, we have considered that the best comparison would be against other off-the-shelf alternatives. We have compared our model with respect to several NARX models available in the Matlab System Identification toolbox \cite{Ljung2012}.

Following this spirit, we have avoided any tuning of our method to the particular benchmarks presented in this section. For instance, by using knowledge about the underlying system of the Silverbox benchmark, we obtained better performance by adding cubic regressors into our NARX model. However, we have not used those custom regressors when reporting the performance of our model.

Regarding the pre-processing step, we have chosen a simple zero-phase second order Butterworth low-pass filter. In this case, the filter parameter $\omega$ represents the cut-off frequency of the filter.

In Figure \ref{fig:prediction} we have plotted the prediction errors on both benchmarks for a number of different models and signal-to-(synthetic)-noise ratios. We have tested the following models:
\begin{itemize}
\item {\footnotesize\textsf{GP-FNARX (SoD)}}: the model presented in this paper using a subset of 512 randomly chosen points (subset of data approximation, SoD \cite{RasWil06}).
\item {\footnotesize\textsf{GP-FNARX (FITC)}}: the model presented in this paper using a FITC sparse GP \cite{QuiRas05} with $M=512$ inducing points chosen at random.
\item {\footnotesize\textsf{GP-NARX (SoD)}}: same as {\footnotesize\textsf{GP-FNARX (SoD)}} but with no pre-filtering of the signals.
\item {\footnotesize\textsf{GP-NARX (FITC)}}: same as {\footnotesize\textsf{GP-FNARX (FITC)}} but with no pre-filtering of the signals.
\item {\small\textsf{* nlarx}}: 3 different NARX models implemented in the Matlab System Identification toolbox \cite{Ljung2012}. Default options. No pre-filtering of the signals.
\item {\small\textsf{* nlarx (filt)}}: same as {\small\textsf{* nlarx}} but using the pre-filtering parameters taken from {\footnotesize\textsf{GP-FNARX (FITC)}} (the computation time for computing those parameters is not taken included in the reported figure).
\end{itemize}
All models have order 10 for the inputs and the outputs.

We observe that the GP-FNARX method with a FITC sparse GP provides the lowest error for all noise levels in both benchmarks. Overall, these results allow us to be optimistic with regards to the prediction capabilities of GP-FNARX models and validate the use of the marginal likelihood as a criterion for model selection in the context of automated pre-processing of the data.

The legend of Figure \ref{fig:prediction} reports the computation times of the experiments when performed on a machine with a 2008 Intel Core i7-920 processor at 2.67 GHz. Although the training time of the GP-FNARX model is higher than the {\small\textsf{wavenet}} and {\small\textsf{treepartition}} models, it is significantly faster than {\small\textsf{sigmoidnet}} yet it provides a lower prediction error.

GP models have a one degree of freedom knob which can be used to trade off accuracy with speed: the number of data points used for SoD or the number of inducing points in FITC. Increasing the number of points is not equivalent to increasing the number of parameters in a parametric model and does not imply any risk of overfitting \cite{RasWil06}. 

%\subsection{Silverbox Benchmark}

%\cite{Espinoza2005}
%Marconato 2012 Identification of the Silverbox Benchmark Using Nonlinear State-Space Models

 %- The marginal likelihood can give very good generalisation error in the test set that requires extrapolation. Much better than what can be achieved by finding the hyperparameters that minimize error on a validation set.

 %- FITC can work much better than a full GP because the assumptions it makes can be more suited to the data.

%\subsection{SYSID09 Wiener-Hammerstein Benchmark}

%%%%%%%%%%%%%%%%%%%%%%%%%%%%%%%%%%%%%%%%%%%%%%%%%%%%%%%%%%%%%%%%%%%%%%%%%%%%%%%%
\section{Conclusions}\label{conc}

We have presented GP-FNARX, an \emph{automated} approach for the identification of nonlinear systems based on autoregressive models implemented with Gaussian processes. Our approach combines the data pre-processing step with the identification task per se. By maximizing the \emph{marginal likelihood} of the probabilistic regression model we obtain a fitting procedure that naturally balances model fit and model complexity.

The Gaussian process model resulting from the identification belongs to the family of \emph{Bayesian nonparametric} models. As such, it does not rely on a rigid parametric functional form and has the capability to adapt to arbitrarily complex data. Moreover, the models can report the uncertainty present in their own predictions. For instance, if the model is used in an operating region which is far from the training data, the model will report higher uncertainty. This feature is useful in applications such as robotics or adaptive control where different control strategies may be appropriate depending on the confidence in the predictions.

One of the reasons why \emph{Bayesian system identification} is not widely used is the lack of methods with a competitive computational cost. We have demonstrated our approach on time series having ${>}10^5$ data points and we have shown how it presents a realistic alternative to parametric nonlinear function approximators. We obtain good performance with a comparable cost and we add the ability to quantify the uncertainty in the model. In addition, our method is easy to apply due to the automation of the pre-processing stage.

%%%%%%%%%%%%%%%%%%%%%%%%%%%%%%%%%%%%%%%%%%%%%%%%%%%%%%%%%%%%%%%%%%%%%%%%%%%%%%%%
%\section*{Acknowledgements}

%%%%%%%%%%%%%%%%%%%%%%%%%%%%%%%%%%%%%%%%%%%%%%%%%%%%%%%%%%%%%%%%%%%%%%%%%%%%%%%%

%\begin{thebibliography}{99}

\bibliographystyle{IEEEtran} %acm, apalike
\bibliography{cdc2013.bib} 

\begin{thebibliography}{10}
\providecommand{\url}[1]{#1}
\csname url@samestyle\endcsname
\providecommand{\newblock}{\relax}
\providecommand{\bibinfo}[2]{#2}
\providecommand{\BIBentrySTDinterwordspacing}{\spaceskip=0pt\relax}
\providecommand{\BIBentryALTinterwordstretchfactor}{4}
\providecommand{\BIBentryALTinterwordspacing}{\spaceskip=\fontdimen2\font plus
\BIBentryALTinterwordstretchfactor\fontdimen3\font minus
  \fontdimen4\font\relax}
\providecommand{\BIBforeignlanguage}[2]{{%
\expandafter\ifx\csname l@#1\endcsname\relax
\typeout{** WARNING: IEEEtran.bst: No hyphenation pattern has been}%
\typeout{** loaded for the language `#1'. Using the pattern for}%
\typeout{** the default language instead.}%
\else
\language=\csname l@#1\endcsname
\fi
#2}}
\providecommand{\BIBdecl}{\relax}
\BIBdecl

\bibitem{Peterka1981}
V.~Peterka, ``Bayesian system identification,'' \emph{Automatica}, vol.~17,
  no.~1, pp. 41 -- 53, 1981.

\bibitem{Barber2006}
D.~Barber and S.~Chiappa, ``Unified inference for variational {B}ayesian linear
  {G}aussian state-space models,'' in \emph{Advances in Neural Information
  Processing Systems (NIPS)}, vol.~20, 2006.

\bibitem{Wills2012}
A.~Wills, T.~B. Sch{\"o}n, F.~Lindsten, and B.~Ninness, ``Estimation of linear
  systems using a {G}ibbs sampler,'' in \emph{Proceedings of the 16th IFAC
  Symposium on System Identification (SYSID)}, 2012.

\bibitem{Ljung1999}
L.~Ljung, \emph{System Identification: Theory for the User}, 2nd~ed.\hskip 1em
  plus 0.5em minus 0.4em\relax Prentice Hall, 1999.

\bibitem{Kocijan2005}
J.~Kocijan, A.~Girard, B.~Banko, and R.~Murray-Smith, ``Dynamic systems
  identification with {G}aussian processes,'' \emph{Mathematical and Computer
  Modelling of Dynamical Systems}, vol.~11, no.~4, pp. 411--424, 2005.

\bibitem{Gutjahr2012}
T.~Gutjahr, H.~Ulmer, and C.~Ament, ``Sparse {G}aussian processes with
  uncertain inputs for multi-step ahead prediction,'' in \emph{Proceedings of
  the 16th IFAC Symposium on System Identification (SYSID)}, vol.~16, 2012, pp.
  107--112.

\bibitem{Ljung2012}
L.~Ljung, \emph{System Identification Toolbox\texttrademark User's Guide
  (R2012b)}.\hskip 1em plus 0.5em minus 0.4em\relax The MathWorks, 2012.

\bibitem{TurDeiRas10}
R.~Turner, M.~P. Deisenroth, and C.~E. Rasmussen, ``State-space inference and
  learning with {G}aussian processes,'' in \emph{13th International Conference
  on Artificial Intelligence and Statistics}, ser. W\&CP, Y.~W. Teh and
  M.~Titterington, Eds., vol.~9, Chia Laguna, Sardinia, Italy, May 13--15 2010,
  pp. 868--875.

\bibitem{Schoen2011}
T.~B. Sch{\"o}n, A.~Wills, and B.~Ninness, ``System identification of nonlinear
  state-space models,'' \emph{Automatica}, vol.~47, no.~1, pp. 39 -- 49, 2011.

\bibitem{Bishop2006}
C.~M. Bishop, \emph{Pattern Recognition and Machine Learning}.\hskip 1em plus
  0.5em minus 0.4em\relax Springer, 2006.

\bibitem{RasWil06}
C.~E. Rasmussen and C.~K.~I. Williams, \emph{Gaussian Processes for Machine
  Learning}.\hskip 1em plus 0.5em minus 0.4em\relax MIT Press, 2006.

\bibitem{QuiRas05}
J.~Qui{\~n}onero-Candela and C.~E. Rasmussen, ``A unifying view of sparse
  approximate {G}aussian process regression,'' \emph{Journal of Machine
  Learning Research}, vol.~6, pp. 1939--1959, 2005.

\bibitem{SneGha06}
E.~Snelson and Z.~Ghahramani, ``Sparse {G}aussian processes using
  pseudo-inputs,'' in \emph{Advances in Neural Information Processing Systems
  (NIPS)}, Y.~Weiss, B.~Sch\"{o}lkopf, and J.~Platt, Eds., Cambridge, MA, 2006,
  pp. 1257--1264.

\bibitem{Schoukens2009}
J.~Schoukens, J.~Suykens, and L.~Ljung, ``Wiener-{H}ammerstein benchmark,'' in
  \emph{Symposium on System Identification (SYSID) Special Session}, vol.~15,
  2009.

\bibitem{Hjalmarsson2012}
H.~Hjalmarsson, C.~R. Rojas, and D.~E. Rivera, ``System identification: A
  {W}iener-{H}ammerstein benchmark,'' \emph{Control Engineering Practice},
  vol.~20, no.~11, pp. 1095 -- 1096, 2012.

\end{thebibliography}

%\end{thebibliography}

\end{document}